\documentclass[runningheads]{llncs}

\usepackage{eccv}

\usepackage{eccvabbrv}

\usepackage{graphicx}
\usepackage{booktabs}

\usepackage{microtype}
\usepackage{graphicx}
\usepackage{booktabs} 

\usepackage{amsmath}
\usepackage{amssymb}
\usepackage{mathtools}
\usepackage{xspace}
\usepackage{adjustbox}

\usepackage{algorithm}
\usepackage{algorithmic}

\newcommand{\lan}{LANISTR\xspace} 

\usepackage[pagebackref,breaklinks,colorlinks,citecolor=eccvblue]{hyperref}
\usepackage{hyperref}

\begin{document}

\title{\lan: Multimodal Learning from Structured and Unstructured Data} 

\titlerunning{LANISTR}

\author{%
  Sayna Ebrahimi, Sercan \"{O}. Ar{\i}k, Yihe Dong, Tomas Pfister\\
\institute{Google Cloud AI Research} 
\texttt{\{saynae, soarik, yihed, tpfister\}@google.com}
}

\authorrunning{S.~Ebrahimi et al.}


\maketitle

\begin{abstract}
  Multimodal large-scale pretraining has shown impressive performance for unstructured data such as language and image. However, a prevalent real-world scenario involves structured data types, tabular and time-series, along with unstructured data.  Such scenarios have been understudied.
  To bridge this gap, we propose LANISTR, an attention-based framework to learn from LANguage, Image, and STRuctured data. The core of LANISTR's methodology is rooted in \textit{masking-based} training applied across both unimodal and multimodal levels. In particular, we introduce a new similarity-based multimodal masking loss that enables it to learn cross-modal relations from large-scale multimodal data with missing modalities.  On two real-world datasets, MIMIC-IV (from healthcare) and Amazon Product Review (from retail), LANISTR demonstrates remarkable improvements, 6.6\% (in AUROC) and 14\% (in accuracy) when fine-tuned with 0.1\% and 0.01\% of labeled data, respectively, compared to the state-of-the-art alternatives. Notably, these improvements are observed even with very high ratio of samples (35.7\% and 99.8\% respectively) not containing all modalities, underlining the robustness of LANISTR to practical missing modality challenge. Our code and models are available at \url{https://github.com/google-research/lanistr}
\end{abstract}

\section{Introduction}\label{sec:intro}
Human brains are natural multimodal learners that can integrate and process information from multiple sources of inputs to form a comprehensive and nuanced understanding of the environment for decision making. 
Inspired by humans' multi-sensory perception, it has also been the overarching goal of machine intelligence to develop multimodal models that can learn meaningful representations from the underlying multimodal data for complex reasoning tasks. 
Multimodal learning have been shown to improve downstream task's performance, robustness, interpretability, and data efficiency~\cite{lpm_survey18,lpm_survey22}.

The literature on multimodal learning has shown striking breakthroughs in modeling unstructured data, specifically vision, language, video and audio modalities~\cite{clip,align,flamingo,coca,beit,videobert,avlnet,merlot,dalle,imagen,latentdiffusion,pali}. In contrast, structured data, including tabular or time-series formats depending on the nature of features (static or time-varying), have been under-explored for multimodal learning despite being the most common data type in the real world~\cite{tabnet,bughin2018notes}.
Numerous real-world applications demonstrate the coexistence of structured data alongside unstructured data, rendering the former a repository of pertinent information. For instance, in healthcare diagnosis prediction, patients' clinical measurements accompany their medical imaging and clinical notes. Similarly, retail demand prediction leverages past sales figures in conjunction with product descriptions, while financial asset price prediction involves past price and volume data coupled with earnings reports. This trend of incorporating structured data into real-world machine learning scenarios is propelled by two interconnected factors. Firstly, cloud-based database management technologies have revolutionized data storage, integration, and manipulation on a massive scale, making it more affordable and convenient. Secondly, the proliferation of multi-sensing technologies, such as wearable devices for humans or intelligent sensors in automobiles and manufacturing facilities, has resulted in the accumulation of high-dimensional time-series data~\cite{lpm_survey22}. Thus, a significant number of real-world machine learning scenarios, initially centered around unstructured data, inevitably encompass relevant structured data which underscores the critical importance of adopting multimodal learning approaches that accommodate structured data

\begin{figure*}[t]
\begin{center}
  \includegraphics[width=0.98\textwidth]{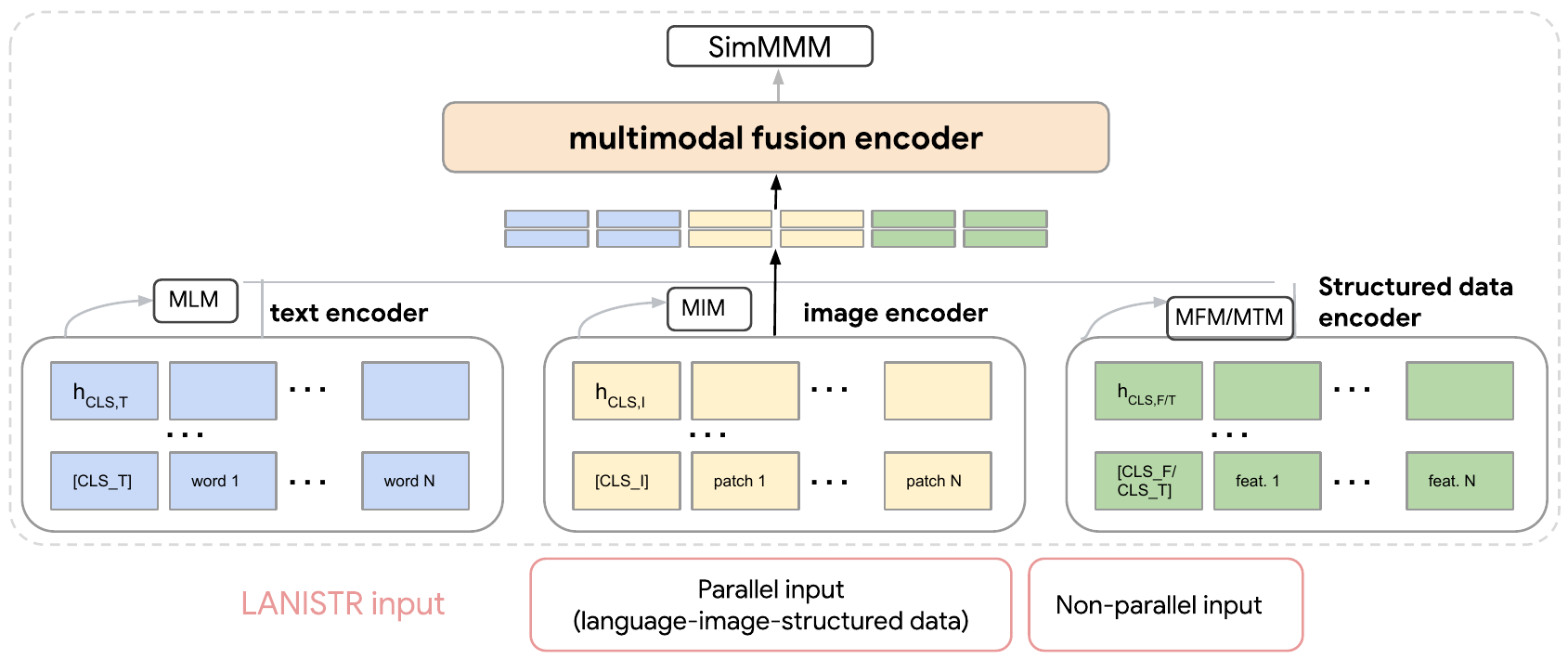}
\end{center}
  \caption{\lan architecture and pretraining objectives. It is composed of modality-specific encoders and a multimodal fusion encoder that combines the concatenated embeddings via cross attention. \lan accepts both parallel (with all modalities present) and non-parallel (data with missing modalities) multimodal data samples.}
\label{fig:lanistr}
\end{figure*}

Unlocking the potential benefits of multimodal learning requires addressing two major challenges that become increasingly prominent as the number of modalities, input size, and data heterogeneity increase. 
First, a fundamental challenge is generalization -- as the input feature dimensionality and heterogeneity increase, deep neural networks can become susceptible to overfitting and suboptimal generalization, particularly when trained on datasets of limited scale. This concern is exacerbated in structured data -- for example, time series often exhibit non-stationary behavior, unlike other more i.i.d. modalities, making it difficult to build well-generalisable models \cite{koopman}. Similarly, tabular data often include numerous features containing minimal information, leading to overfitting to spurious correlations~\cite{tabnet}.
Second, modality missingness becomes a more prominent issue when dealing with multimodal data beyond two modalities, with it being likely that there are samples that miss at least one modality. 
To the best of our knowledge, a systematic study on learning from unstructured and structured data that addresses these challenges remains absent from current literature.

Consequently, we pose the following question: \textit{Given the aforementioned challenging differences between structured and unstructured data, does it empower the overall representation when we learn them together?} 
We hypothesize the answer is \textit{yes} and set the basis of our work to answer the following question: \textit{How can we learn two seemingly very different data types together in a multimodal fashion with a unified architecture and unique pretraining strategies that resemble the nature of a dataset with structured and unstructured modalities?}

In this work, we propose \lan, a novel framework for multimodal learning with unstructured (vision and language) and structured data (tabular and/or time series). 
\lan learns a unified representation through joint pretraining on all the available data with significant missing modalities. 
\lan leverages unimodal masking pretraining while encompassing cross-modal relationships through a \textit{similarity-based multimodal masking} objective. 
As depicted in Fig. \ref{fig:lanistr}, the \lan model processes input raw multimodal data, which can be either parallel (without any missing modality) or non-parallel (with some modalities being missing) and encodes them through modality-specific encoders. The resulting embeddings are then concatenated and fed into the proposed multimodal fusion encoder. This fusion encoder, implemented based on an attention-based architecture, conducts cross-attention interactions among the projected unimodal image, text, and structured data (tabular and/or time series) representations, effectively fusing all modalities into a unified framework. 
Our contributions and key demonstrations include:
\begin{itemize}
    \item In multimodal pretraining, multimodal objectives can bring significant gains beyond unimodal ones, as they can encourage better joint learning. However, extending conventional pretraining strategies from unstructured data, like contrastive pretraining, to multiple modalities alongside structured data is challenging. To address this, we propose a framework exclusively built upon unimodal and multimodal masking techniques for pretraining.
    \item We show that utilizing large scale unlabeled data can bring significant gains for multimodal learning even in the presence of missing modalities for most data samples, a commonly-observed real-world scenario. Our proposed similarity-based multimodal masking pretraining objective adeptly addresses the missingness challenge, proving highly effective for this purpose.
    \item Our findings highlight self-supervised pretraining's effectiveness in superior out-of-distribution generalization, even with scarce and dissimilar labeled tuning data -- a common situation in domains such as retail and healthcare, particularly with structured data. We show \lan's capability to be pretrained on a specific shopping category of the Amazon Product Review data (\textit{Office Products}), achieving a remarkable absolute 23\% accuracy boost when fine-tuned on a distinct category such as \textit{Fashion Products}. This performance boost is achieved using a mere 0.01\% of data (512 labeled samples).    
\end{itemize}

\section{Related work} 

\noindent \textbf{Self-supervised multimodal learning.} Self-supervised multimodal learning can be considered under three
categories based on their objective: instance discrimination-based, clustering-based, and masked prediction-based. \textbf{Instance discrimination-based} approaches are based on contrastive or matching prediction. For contrastive learning, samples from two modalities are selected as positive/negative pairs, and the model is trained to distinguish the two using a contrastive objective~\cite{clip,cmc,avts,mmv,vatt}. CLIP~\cite{clip}, pretrained on $\sim$400M of image-text pairs, achieves impressive zero-shot performance and has been successfully extended to other modalities, e.g. AudioCLIP~\cite{audioclip} and VideoCLIP~\cite{videoclip}, however, obtaining pairs/triplets of modalities is not always feasible. Also, as the number of modalities and dataset size increase, it becomes computationally more expensive to train different modalities in a contrastive way. Matching prediction aims to predict whether a pair of samples from two modalities are matched or not, and has been used for audio-visual correspondence~\cite{avc1, avc2} or image-text matching (ITM)~\cite{uniter}, also adopted by~\cite{flava,visualbert}. 
\cite{albef} use both ITM and contrastive learning together to fuse image and text modalities through cross attention. \textbf{Clustering methods}~\cite{xdc,dmc,uhubert} learn the underlying data structure through the iterative process of predicting the cluster assignments in the encoded representation, and using pseudo labels to update the feature representations. Multimodal cluster assignments allow different modalities to have different assignments to increase diversity but the paired modalities might not be perfectly matched and it is hard to know apriori the optimal flexibility. For noisy paired datasets, clustering approaches can alleviate the issue of false positives and hard negatives that contrastive learning suffers from, however, there are still challenges including scalability, sensitivity to parameter initialization, the choice of clustering algorithm, and determining the optimal number of clusters.
\textbf{Masked prediction-based methods} can be either performed with an auto-encoding (similar to BERT~\cite{bert}) or an auto-regressive approach (similar to GPT~\cite{gpt}). Auto-encoding masked predictors pretrain models by predicting randomly masked pieces in the input, encouraging to learn rich semantic features. It was first introduced for text data~\cite{bert} and is widely used for multimodal tasks as well, for which, the masked signal is predicted conditioned on other modalities, encouraging understanding of the cross-modal interactions. Intra-modal masking can also be used, predicting the masked information contained with the same modality~\cite{videobert, unifiedio, beit3}. Auto-regressive masked predictors, popular in computer vision~\cite{pixelcnn} and NLP~\cite{gpt}, aim to predict the next masked token given the previous ones. However, they have been adopted less for multimodal learning compared to auto-encoding~\cite{simvlm,ssmml} as auto-encoding masked predictors can be easier and faster to train.
There are multimodal learning approaches that combine the auto-encoding and auto-regressive masked predictions -- e.g., Omni-perception Pretrainer~\cite{opt} learns image-text-audio multimodal representations by auto-encoding masking at token level for vision and language, and auto-regression masking at the modality level using modality-specific decoders. \lan leverages modality-specific auto-encoding masking with the randomly masked information in each modality using a reconstruction loss. Beyond these, we introduce a novel multimodal masking objective that aims to overcome the missing modality challenge by maximizing the similarities between masked and unmasked data representations. 

\noindent \textbf{Learning with unstructured and structured data.} Recent impressive success of large language models (LLMs) have led to the idea of converting structured data to unstructured text to allow processing them with LLMs~\cite{tabllm} using simple approaches such as feature concatenation, or more complicated approaches such as table-to-text generation~\cite{tabletotext_pivot,tabletotext_neural}. Training table-to-text generation models requires paired table and text data, and is computational expensive. Moreover, for multimodal datasets with a large number of categorical features, it is prohibitive to concatenate the tabular features with language token sequences as the sequence length is fixed. Furthermore, for time-series data with only numerical values, conversion to text might be quite suboptimal due to the distribution mismatch of the such sequences with text data. \lan overcomes these challenges by having a modality-specific encoder in its architecture for tabular or time series, allowing for proper representation encoding for all the modalities separately. 
For this multimodal learning scenario, one proposed solution is AutoGluon~\cite{autogluon} that can learn from labeled text, image, and tabular data with a fusion model based on MLP or Transformer.
There is also previous research specifically in the healthcare domain ~\cite{zhang2020combining,arnaud2020deep,medfuse}, often with architectures that are not attention-based such as convolutional, MLP or LSTM-based, with multimodal learning being based on a simple late fusion~\cite{zhang2020combining} or embeddings fused with an LSTM ~\cite{medfuse}.

\noindent \textbf{Multimodal learning with missing modalities.} Learning with non-parallel data, \ie data with missing modality, reflects the common real-world scenario of the coexistence of some parallel data and a larger amount of non-parallel data. 
Since Transformers can be sensitive to missing modalities~\cite{Ma_2022_CVPR}, self-supervised learning methods for dealing with mixed-parallel data usually apply separate pretext tasks for parallel and non-parallel data in a multi-task manner with masked prediction being one task~\cite{flava,unimo,videobert}. FLAVA~\cite{flava} employs masked image and language modeling for image-only and text-only data via modality-specific encoders, while utilizing masked multimodal modeling and contrastive learning over paired data with a multimodal Transformer. UNIMO~\cite{unimo} applies masked image modeling to image-only data, masked language modeling, and sequence-to-sequence generation~\cite{unilm} to language-only data. For this challenge, in \lan, pretraining unimodal encoders with masked signal modeling objectives, our approach is based on random masking input modalities in parallel data triplets to enforce similar embeddings to non-masked inputs.  

\section{\lan: a framework for LANguage, Image, and STRuctured data}
In this section, we introduce our proposed framework, \lan, for multimodal learning from structured and unstructured data. We present how \lan is pretrained on unlabeled data using unimodal and multimodal masking-based objectives and we provide insights on how its pretraining objectives are designed to help with missing modality. Lastly, we explain how a pretrained \lan can be used for learning different downstream tasks. Note that specific details and hyperparameters are provided in the Appendix. 

\subsection{Model architecture}
Fig.~\ref{fig:lanistr} overviews the model architecture of \lan, which is composed of modality-specific encoders and a multimodal encoder-decoder module as the fusion mechanism. 
First, raw inputs are encoded with a language encoder, an image encoder, and a structured data encoder. Depending on the dataset, we can have two separate structured data encoders, one for tabular data and one for time-series. These modality-specific encoders are all chosen to be attention-based architectures.

After the embeddings are obtained from the inputs of each modality, they are concatenated and fed into a multimodal fusion encoder module. The hidden state vectors obtained by encoding the inputs are projected using modality-specific encoders with a single layer projection head and the results are concatenated together to feed them into the multimodal fusion module. 

One bottleneck for machine learning with multimodal data is extracting meaningful representations that reflect cross-modal interactions between individual modalities.
As the fusion encoder, we adopt a cross-attention architecture, based on a Transformer architecture, to better capture cross-modal relationships.

\subsection{Pretraining objectives}
\lan is pretrained with two types of objectives (i) unimodal masking losses and (ii) similarity-based multimodal masking loss, that both contribute to better learning of meaningful representations of multimodal data. These are described in detail in the following sections.

\subsubsection{Unimodal self-supervised learning}
We use masked \textit{signal} modeling as a general self-supervised learning strategy for all the unimodal encoders in \lan. This allows utilizing non-parallel data for unimodal encoders, as masked inputs are fed to encoders and a form of reconstruction or prediction task can be used for training. We describe four types of unimodal masking losses for language, image, tabular, and time series modalities:  

\noindent \textbf{Masked Language Modeling} $(\mathcal{L}_\text{MLM})$~\cite{bert,roberta} and its auto-regressive variants~\cite{gpt,gpt2,gpt3} are the most dominant self-supervised learning strategies for LLMs. Following~\cite{bert}, we integrate a classifier head on top of the text encoder (BERT~\cite{bert}), to perform the task of predicting masked tokens out of the entire vocabulary given the unmasked tokens. 

\noindent \textbf{Masked Image Modeling $(\mathcal{L}_\text{MIM})$.} 
As the image encoder, we adopt an attention-based architecture (ViT-B/16~\cite{vit}) and employ image masking based pretraining, as also used in SimMIM~\cite{simmim}. For this pretraining, the task is to reconstruct raw pixels of masked image patches given the rest of the image. We use a linear layer on top of the latent feature representation of the image encoder for image reconstruction and train it with an $l_1$ loss.

\noindent \textbf{Masked Feature Modeling $(\mathcal{L}_\text{MFM})$.} We adopt TabNet~\cite{tabnet} for encoding tabular (time-invariant structured data) features and follow its self-supervised masking strategy to pretrain the tabular encoder where the task is to reconstruct missing tabular feature given the visible columns. Following~\cite{tabnet}, we use a decoder on top of the encoder with feature Transformers, followed by fully-connected layers at each decision step. The decoder is only used during pretraining and is discarded during the supervised fine-tuning stage. The outputs from the decoder are averaged to obtain the reconstructed features. 

\noindent \textbf{Masked Time series Modeling $(\mathcal{L}_\text{MTM})$.} We use a conventional attention-based Transformer as the time series encoder and train it with the standard self-supervised masking modeling objective by defining the task of regressing to masked values. In particular, we define a binary noise mask for each data point where on average we set 15\% of each column of data (corresponding to a single variable in the multivariate time series) to zero. We follow~\cite{mvts} in using a geometric distribution for masked segments to prevent the model from trivially predicting the missing values by replacing with the immediately preceding or succeeding values, or their averaged value. We use a linear layer on top of the encoder's final embeddings, output a vector of equal size with the input and compute the mean squared error loss for the masked values for supervision. Hence, this is different from the conventional denoising used in autoencoders, where the entire input is injected with Gaussian noise and is reconstructed as a whole. 

\begin{figure}[t]
\centering
  \includegraphics[width=0.65\columnwidth]{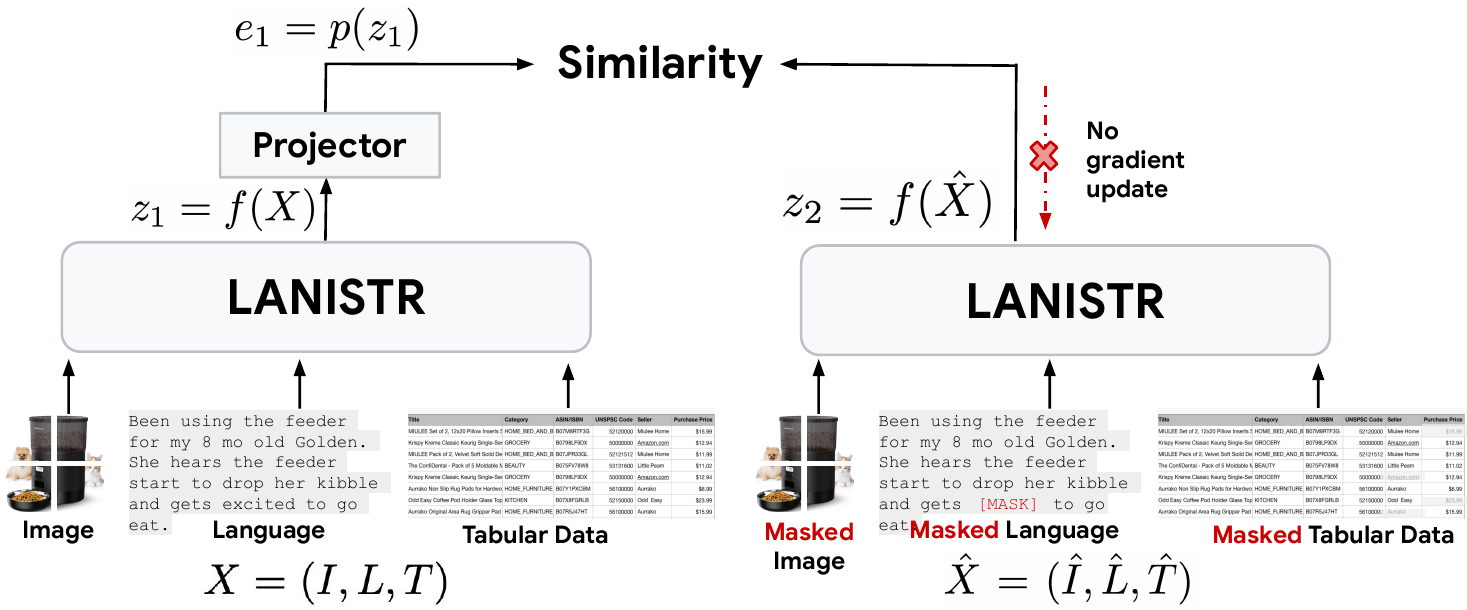}
  \caption{Illustration of similarity-based multimodal masking in \lan based on the objective defined between the multimodal input and its masked version.}
\label{fig:simmmm}
\end{figure}
\subsubsection{Multimodal self-supervised learning} Prior work on multimodal learning have focused on \textit{reconstructing} one modality (\eg text~\cite{visualbert}) or both image and text modalities~\cite{flava} from the masked multimodal inputs. However, in this work, we propose a novel masked multimodal learning loss that maximizes the similarities between masked and unmasked multimodal data representations. This objective resembles of an idea that was originated from the Siamese networks~\cite{simsiam} where the goal is to maximize the similarity between two augmented versions of an image. However, in our framework, the goal is to maximize the similarity between the embeddings generated by a masked and a non-masked input.

Assume the input data samples are in the form $X = (I, L, T)$ where $I$, $L$, and $T$ represent image, language, and time series/tabular modality inputs. We create masked views of the data triplets denoted as $\hat{X}=(\hat{I}, \hat{L}, \hat{T})$ by randomly masking a portion of the input, i.e., either removing some image patches, replacing some sub-words in the text with [\texttt{MASK}] token, masking some values in columns in the tabular data, or removing some timestamps in a series of time events. The architecture receives $(I, L, T)$ and $(\hat{I}, \hat{L}, \hat{T})$ as two inputs which are processed through the unimodal encoders followed by the multimodal fusion encoder which unlike the unimodal encoders shares weights across different modalities. Fig.~\ref{fig:simmmm} shows that $f$, which represents the entire \lan architecture, is followed by a projector $p$ which takes in the output of the multimodal encoder, denoted as $z_1 = f(X)$, and projects it to a final embedding \ie $e_1 = p(z_1)$. We define the output embedding of a masked input as $z_2 = f(\hat{X})$ and minimize the negative cosine similarity between $e_1 $ and $z_2$ as 
\begin{equation}
\mathcal{D}(e_1, z_2) = - \frac{e_1}{||e_1||_2} \cdot \frac{z_2}{||z_2||_2},
\end{equation}
where $||\cdot||_2$ is $l_2$-norm. Inspired by ~\cite{byol,simsiam}, we propose total masking multimodal loss as a symmetric function as follows:
\begin{equation}
\mathcal{L}_{\text{SimMMM}} = \mathcal{D}(e_1, z_2) + \mathcal{D}(e_2, z_1).
\end{equation}
This objective encourages the model to learn cross-modal relations such that the cosine similarities between the embeddings of masked and non-masked data samples are maximized. We have observed this objective to be more effective in learning cross-modal relationships and to bring more robustness to missing modalities, whereas the reconstruction-based masking objectives used for unimodal encoders encourages learning modality-specific features.
We follow the ``stop gradient'' operation introduced in ~\cite{simsiam} in our implementation which prevents the encoder on $\hat{X}$ from receiving gradients from $z_2$ in the first term while receiving gradients from $e_2$ in the second term (and vice versa for $X$). \cite{simsiam} shows that without applying ``stop gradient'' operation, the optimizer can lead to a degenerate solution and reaches the minimum possible loss of -1, while adding it yields smooth convergence.

By combining all unimodal masking losses and the multimodal similarity-based masking loss, we obtain the full objective function for \lan pretraining as:
\begin{equation}
\label{eq:total_loss}
\mathcal{L}_{\text{\lan}} = \lambda_1 \mathcal{L}_{\text{MLM}} + \lambda_2 \mathcal{L}_{\text{MIM}} + \lambda_3 \mathcal{L}_{\text{MFM}}  + \nonumber \lambda_4 \mathcal{L}_{\text{MTM}} + \lambda_5 \mathcal{L}_{\text{SimMMM}},
\end{equation}
where $\lambda_i$ with $i=\{1, \cdots, 5\}$ are hyperparameters that determine the effect of each loss component during pretraining. Algorithm~\ref{alg:lanistr} shows the pseudocode for self-supervised pretraining with \lan. We discuss selection of hyperparameters for \lan in the Appendix.

\subsection{Fine-tuning \lan} 
For most real-world scenarios, the amount of labeled data available for fine-tuning would be much smaller than the amount of unlabeled data available for pretraining. Thus, mechanisms to bring robustness against overfitting becomes of vital importance, which is addressed in \lan by controlling frozen vs. trainable layers.

After pretraining, we use pretrained weights to initialize both the unimodal encoders and the multimodal encoder. We integrate an MLP classification module with the multimodal encoder for the downstream task. We propose keeping the unimodal encoders in a frozen state while concentrating on training the multimodal encoder and the classification module.\footnote{This accounts for training approximately 15\% of the entire \lan architecture with more than 270M parameters for the selected hyperparameters used in experiments (see Appendix for more details).} It's worth noting that \lan's versatility can be extended to other tasks, such as regression or retrieval, by incorporating suitable heads and objective functions provided labeled data is accessible.


\section{Experimental Setup}\label{sec:experiments}
In this section, we provide details about the experimental settings, datasets, tasks, and the implementation for evaluating pretraining strategies in \lan. 


\subsection{Datasets} For our experiments, we focus on two large-scale real-world datasets consisting image, text, and structured data  (either as tabular or time series) modalities, described below.

\noindent\textbf{MIMIC-IV (v2.2)} or Medical Information Mart for Intensive Care~\cite{mimic4} is a popular public medical dataset for clinical prediction tasks. We consider the binary task of predicting in-hospital mortality after the first 48-hours of ICU stays. We use clinical time series data collected during this period, clinical notes by the medical team, and the last chest X-ray image taken in the first 48-hour time window for the image modality. For time-series preprocessing, we follow standard benchmarks such as~\cite{medfuse,harutyunyan2019multitask}; and for image and text modalities we follow common practice of image transformations and text preprocessing schemes used in masked image~\cite{simmim} and language~\cite{bert} modeling techniques. The pretraining dataset has 3,680,784 samples from which 1,315,592 miss at least one modality (35.7\% missingness ratio). For fine-tuning, we have only 5923 labeled samples from which 5298 are used for training, while 8 and 617 are used for validation and test sets, respectively. Data preprocessing and detailed statistics are given in the appendix.

\noindent\textbf{Amazon review data (2018)} \cite{apr} contains reviews and metadata spanning 1996-2018 across diverse product categories. The objective is to predict the star rating (out of 5) a product receives. Our experiment employs \textit{Office Products}, \textit{Fashion}, and \textit{Beauty} categories. Pretraining utilizes $\text{5,581,312}$ samples from the \textit{Office Product} category, whereas fine-tuning focuses on a parallel subset of 512 training samples from \textit{Fashion} and \textit{Beauty}, with a validation and test set of 128 and 256 samples, respectively. For parallel data, triplets encompass image, text, and tabular features. Product images include seller or user-provided visuals, truncated text summaries, and full reviews limited to 512 characters. Tabular features encompass product ID, reviewer ID, review verification status, year, review ratings count, and timestamp. Data preprocessing and detailed statistics are provided in the appendix. Our fine-tuning categories aim to evaluate generalization capacity, leveraging a substantial unlabeled dataset for learning from a significantly smaller dissimilar labeled subset.

\subsection{Baselines}\label{sec:baselines}


In this section, we overview the baselines that we compare \lan against. While \lan can be used for multimodal settings with both tabular and time series, to the best of our knowledge there is no architecture and pretraining strategy that is specifically designed for image, text, and both types of structured data. Hence, we consider popular fusion methods to be able to exploit all modalities and modify the state-of-the-art dual modality baselines from vision and language learning, by fusing structured data into them as text to establish a new baseline for image, text, and structured data.  

\noindent \textbf{LateFusion (image+text+tabular/time series)} is a simple fusion mechanism where we use modality-specific encoders followed by a projection layer for each encoder before concatenating all their embeddings and feed them to a classifier head. We train all the encoders, the projection layers, and the classifier head end-to-end using only the parallel labeled data. We use off-the-shelf pretrained ViT-B/16 image encoder and BERT-base uncased text encoder for initialization. 

\noindent \textbf{AutoGluon 
\cite{multimodalautogluon}}\footnote{https://auto.gluon.ai/} is 
is similar to our late fusion baseline which enables training a multimodal model for labeled image, text, and tabular data (not time-series) by end-to-end training a ViT-B/16 image encoder, a BERT text encoder, and an MLP tabular data encoder that are concatenated and fed to an MLP-style or a vanilla Transformer fusion encoder. AutoGluon can handle missing image modality only by replacing the pixels with zeros.

\noindent \textbf{FLAVA~\cite{flava} (image+text) } is a foundation model for vision and language that can be trained on both paired and unpaired data using unimodal masking losses, CLIP-style~\cite{clip} global contrastive loss, image-text matching loss, and masked multimodal loss where for the latter the task is to predict the masked patch in the image similar to BeiT~\cite{beit} and word vocabulary index of the masked text tokens. It is composed of BERT and ViT for text and image encoders which are then fused using a ViT multimodal encoder.
We use only image and text modalities for this baseline as it cannot use tabular or time series modalities.

\noindent \textbf{CoCa~\cite{coca} (image+text)} is an image-text encoder-decoder foundation model which is jointly trained with contrastive loss and captioning loss. We use the released checkpoint by OpenCLIP library~\cite{openclip}~\footnote{The checkpoint is available on HuggingFace library as \texttt{laion/mscoco\_finetuned\_CoCa-ViT-L-14-laion2B-s13B-b90k}} and fine tuned it on image and text data only. The contrastive loss weight is set as 0 for finetuning as recommended by OpenCLIP. 

\noindent \textbf{ALBEF~\cite{albef}} is a strong vision and language model that we use as is (without tabular modality) as well as with tabular modality where tabular data is fused as text to the model when available and time series modality is discarded. ALBEF pretrains the text encoder using a masking loss before aligning image and text modalities using an image text matching loss and a MoCo-style image-text contrastive loss~\cite{moco}. It consists of a ViT and BERT for image and text encoders where their features are fused together through cross attention at each layer of a multimodal encoder which has an architecture similar to the last 6 layers of BERT. 
    

\noindent \textbf{MedFuse~\cite{medfuse} (image+time series)} employs a simple LSTM-based fusion mechanism with independently pretrained modality-specific encoders.  Specifically, it uses ResNet-34 for images (pretrained for 14-way disease classification on unpaired chest X-rays) and an LSTM for time series (pretrained on unpaired EHR data for in-hospital mortality prediction).  After pretraining, the classifiers are removed and the encoders, projection layers, and LSTM fusion module are fine-tuned on paired image and time series data for in-hospital mortality prediction. While MedFuse utilizes unpaired data, its pretraining focuses on separate tasks for each modality, limiting the learning of cross-modal relationships. We evaluate MedFuse using their publicly available package on the MIMIC-IV-v2.2 dataset with splits consistent with \lan and other baselines
    
\noindent \textbf{Tab2Txt} is employed on top of other baselines to feed tabular data as text to their models, as in \cite{dinh2022lift,narayan2022foundation}. It is based on converting the tabular features into a string format and prepending them to the text input. 
This baseline fundamentally suffers from the limitation that the pretraining data coverage of text encoders for structured data, especially with numerical features, would be often insufficient, resulting in suboptimal learning for tabular or time-series data. Moreover, limited context length of text encoders often limits the applicability of this approach to large-scale real-world tabular or time-series data (and even when they fit in the context length, it can be suboptimal for the text encoder models \cite{liu2023lost}).
We focus on this baseline to highlight the importance of employing a separate tabular and time-series encoder, considering it on top of ALBEF and \lan.

\section{Results and Discussions}
We first show evaluations for \lan compared to the key baselines. Then, we present ablation studies to demonstrate the effect of key components of \lan. 

\begin{table}[t]
\centering
\scriptsize 
\caption{Results for MIMIC-IV dataset. Results for MedFuse, LateFusion and \lan are averaged over three runs.
}\label{tab:mimic}
\begin{tabular}{@{}lc@{}}
\toprule
Method/Category & AUROC \\ \midrule
CoCa &  38.45 \\
FLAVA &  77.54 \\
MedFuse & 78.12 $\pm$ 2.79 \\
LateFusion & 80.79 $\pm$ 1.12 \\
\textbf{\lan}, no pretrain & 80.87 $\pm$ 2.56 \\
\textbf{\lan} & \textbf{87.37 $\pm$ 1.28} \\ \bottomrule
\end{tabular}
\end{table}

\subsection{Results on MIMIC-IV}
Table~\ref{tab:mimic} shows comparison of \lan against baselines on MIMIC-IV dataset. CoCa is only finetuned with image and text data and despite having 638.45M params (2x larger than \lan) only achieves 38.45\% in AUROC. While CoCa has shown excellent performance in text generation tasks, its performance on mortality prediction is low using text and image modalities only. Another potential reason could be that the publicly available checkpoint for this dataset is not as optimal as the original unreleased model. FLAVA, although finetuned with text and image only, is better at discriminative tasks compared to CoCa and yields 77.54\% AUROC. MedFuse, as the state-of-the-art multimodal (time series and image) model that is specifically designed for this dataset, achieves 78.12\% AUROC; while late fusion with Transformer-based encoders achieves 80.79\% AUROC. This shows the effect of using more advanced encoders and more modalities in LateFusion compared to the ResNet and LSTM encoders used in MedFuse which slightly surpasses the effect of pretraining with unpaired data in MedFuse. On the other hand, \lan without pretraining achieves 80.87\%, slightly better than LateFusion while pretraining \lan with unlabeled data improves the performance to 87.37\% of AUROC, which renders is significantly better than all others.

\subsection{Results on Amazon Product Review}
Table~\ref{tab:amazon} compares \lan with AutoGluon, ALBEF, and LateFusion baselines on the two categories of the Amazon dataset. Among the baselines, ALBEF is the only one that utilizes pretraining image and text data. We present two sets of results for this method -- one in its original form with image and text modalities and the second one when the tabular data are included in the text modality, as previously defined as Tab2Txt baseline. For AutoGluon, we use two possible fusion mechanisms provided in its package, \ie MLP and Transformer-based fusion. In the experiment for the \textit{Beauty} category, \lan is able to achieve 76.27\% average accuracy, and outperforms all the baselines by a large margin. AutoGluon with $\sim$200M parameters achieves 55.34\% and 61.59\% accuracy using the MLP and Transformer fusion mechanisms, respectively. The LateFusion baseline, which uses TabNet as the tabular encoder and a small MLP fusion mechanism, achieves 62.47\% accuracy. This highlights the importance of encoding tabular features with an attention-based encoder instead of an MLP as in AutoGluon. ALBEF, in its original form, achieves 56.34\% accuracy which is mainly due to leveraging the unlabeled data despite not having access to the tabular information. When we feed categorical features represented as text to ALBEF, the accuracy is degraded, showing the importance of reviews over tabular features for this task as prepending tabular features results in a shorter text token sequence because the total maximum input size is limited. \lan without any pretraining still achieves a reasonable accuracy (65.43\%) even though the downstream task data for the \textit{Beauty} category is substantially different from the pretraining data on the \textit{Office Products} category. \lan + Tab2Txt achieves lower accuracy (59.23\%) compared to \lan, which demonstrates the importance of processing unstructured and structured data separately. 

In the experiment on the \textit{Fashion} category, \lan outperforms AutoGluon by a large margin, with an absolute difference in accuracy of up to $\sim$24\%. This is mainly attributed to improved multimodal learning architecture and pretraining methods of \lan.
On the other hand, LateFusion achieves 65.83\% accuracy, which is higher than the accuracy of \lan without pretraining, but much lower than the accuracy of \lan with pretraining. 
Although the high capacity of \lan might suffer from poorer generalization when trained with a small dataset size of 512 samples, we observe that with proposed multimodal pretraining, the generalization is significantly improved and significant outperformance is obtained.
Similar to the results on the \textit{Beauty} category, converting the tabular input to text in \lan achieves a lower accuracy of 48.21\%, highlight the importance of separate representation learning via structured data encoders and proposed pretraining objectives of \lan.

\begin{table}[t]
\centering
\scriptsize 
\caption{Results for Amazon Review dataset. AutoGluon~\cite{multimodalautogluon} encodes tabular data using an MLP while for ALBEF~\cite{albef} we feed tabular features as additional text. Methods that can use unlabeled data (\lan and ALBEF) are pretrained on \textit{Office Products} category first. Results are averaged over five runs.}
\begin{tabular}{@{}lccc@{}}
\toprule
Method/Category & \textit{Beauty} & \textit{Fashion} \\ \midrule
AutoGluon-MLP & 55.34~$\pm$~3.55 & 50.39~$\pm$~1.70 \\
AutoGluon-TF & 61.59~$\pm$~4.50 & 46.10~$\pm$~3.92 \\
LateFusion & 62.47~$\pm$~3.32 & 65.83~$\pm$~6.85 \\
ALBEF, Tab2Txt & 43.51~$\pm$~2.91 & 43.23~$\pm$~3.56 \\
ALBEF & 56.34~$\pm$~2.09 & 55.78~$\pm$~2.16 \\
\textbf{\lan}, Tab2Txt & 59.23~$\pm$~3.76 & 48.21~$\pm$~4.62 \\
\textbf{\lan}, no pretrain & 65.43~$\pm$~7.13 & 52.07~$\pm$~5.66 \\
\textbf{\lan} & \textbf{76.27~$\pm$~3.17} & \textbf{75.15~$\pm$~1.20} \\ 
\bottomrule
\end{tabular}
\label{tab:amazon}
\end{table}


\subsection{Ablation studies}
Table~\ref{tab:ablation_mimic} shows the ablation studies for different objective functions in \lan as well as on the employment of different modalities on MIMIC-IV dataset, described below. 

\subsubsection{Gains from different modalities.} When a particular modality is not used for ablation, its associated masking loss is also removed from pretraining. Ablating the text modality results in the lowest AUROC of 70.29\%, followed by the image modality with 72.78\% and the time series modality with 79.89\%. This highlights the importance of information in each modality of this particular dataset, as well as how \lan leverages each modality when available.
 
\subsubsection{Unimodal vs. Multimodal self-supervised learning.} In pretraining objectives, omission of SimMMM and MLM leads to the most significant performance decline, resulting in 80.43\% and 80.89\% respectively. Ablating MTM has the least impact, followed by MIM. 

\begin{table}[t]
    \centering
    \caption{Ablation study for modalities and objective functions in \lan in the presence of different modalities in the MIMIC-IV dataset.} \label{tab:ablation_mimic}
    \begin{adjustbox}{width=0.8\textwidth}    
    \begin{tabular}{@{}c|ccc|cccc|c|c@{}}
        \toprule
        Ablation & \begin{tabular}[c]{@{}l@{}}w/o\\ time\end{tabular} & \begin{tabular}[c]{@{}c@{}}w/o\\ image\end{tabular} & \begin{tabular}[c]{@{}c@{}}w/o\\ text\end{tabular} & \begin{tabular}[c]{@{}c@{}}w/o\\ $\mathcal{L}_\text{MTM}$\end{tabular} & \begin{tabular}[c]{@{}c@{}}w/o\\ $\mathcal{L}_\text{MIM}$\end{tabular}& \begin{tabular}[c]{@{}c@{}}w/o\\ $\mathcal{L}_\text{MLM}$\end{tabular}   & \begin{tabular}[c]{@{}c@{}}w/o\\ $\mathcal{L}_\text{SimMIM}$\end{tabular}  & \parbox{1.5cm}{\centering w/o non-\\parallel data} & \lan \\
        \midrule
        AUROC & 79.89 & 72.78 & 70.29 & 83.41 & 82.23 & 80.89 & 80.43 & 79.87 & 87.37 \\
        \bottomrule
    \end{tabular} 
    \vspace{-10pt}
    \end{adjustbox}
\end{table}

\subsubsection{Learning from data with partially-available modalities.} 
Excluding non-parallel data results in a 6.34\% AUROC reduction compared to \lan's performance. This implies that \lan effectively forges cross-modal relationships and uses the absence of modalities to its advantage rather than being hindered by it.

\begin{table}[t]
\centering
\small
\caption{Effect of pretraining dataset size on downstream task in MIMIC-IV.}
\begin{adjustbox}{width=0.5\textwidth}
\begin{tabular}{@{}cccccc@{}}
\toprule
\% Unlabeled Data & 0\% & 25\% & 50\% & 75\% & 100\% \\ \midrule
AUROC (\%) & 80.87 & 81.90 & 83.60 & 85.90 & 87.37 \\
\bottomrule
\end{tabular}
\end{adjustbox}
\label{tab:pretraining_data}
\end{table}

\subsubsection{Effect of pretraining dataset size} 
Table~\ref{tab:pretraining_data} shows an ablation on pretraining dataset size where increasing its size improves downstream task performance. This demonstrates \lan's ability to consistently leverage unlabeled data when it is fine-tuned on merely 0.1\% labeled data.  


\section{Conclusion}
We present \lan, a novel framework for language, image, and structured data, utilizing unimodal and multimodal masking strategies for pretraining. Our innovative similarity-based multimodal masking objective addresses the challenge of missing modality in large-scale unlabeled data, a prevalent issue in real-world multimodal datasets. Demonstrated on real-world retail (Amazon Product Review) and healthcare (MIMIC-IV) datasets, \lan showcases remarkable performance improvements over existing methods. Notably, \lan achieves impressive out-of-distribution results despite limited labeled data. 

%
%
\clearpage
\bibliographystyle{splncs04}
\bibliography{main}

\newpage
\appendix
\onecolumn
{
\begin{center}
\Large 
\textbf{LANISTR: Multimodal Learning from Structured and Unstructured Data}\\(Supplementary Materials)
\par
\end{center}
\vspace{2pt}
}

\section{Hyper-parameters in \lan}
We present all the hyper-parameters used in \lan architecture during pretraining and fine-tuning stages.

  \begin{table*}[!htb]
    \caption{Hyper-parameters used in our pretraining and fine-tuning experiments on MIMIC-IV (left) and Amazon Review (right) datasets.}
    \label{tab:hyperparams}
    \begin{minipage}{.5\linewidth}
    \centering
    \tiny
    \caption*{\small MIMIC-IV dataset}
            \begin{tabular}{|ll|}
            \hline
            \multicolumn{1}{|l|}{Hyper-parameter} & Value \\ \hline
            \multicolumn{2}{|c|}{\textbf{Text Encoder}}           \\ \hline
            \multicolumn{1}{|l|}{HuggingFace model name}   & bert-base-uncased   \\ \hline
            \multicolumn{1}{|l|}{Number of heads}           &    12   \\ \hline
            \multicolumn{1}{|l|}{Number of layers}          &    12   \\ \hline
            \multicolumn{1}{|l|}{Hidden size}               &  768      \\ \hline
            \multicolumn{1}{|l|}{Intermediate size}         &   3072    \\ \hline
            \multicolumn{1}{|l|}{Projection size}         &  768     \\ \hline
            \multicolumn{1}{|l|}{Vocab size}         &  30522   \\ \hline
            \multicolumn{1}{|l|}{Maximum sequence length}         &  512   \\ \hline
            \multicolumn{1}{|l|}{Masking ratio}               &   0.15    \\ \hline
            \multicolumn{2}{|c|}{\textbf{Image Encoder}}          \\ \hline
            \multicolumn{1}{|l|}{HuggingFace model name}           &   google/vit-base-patch16-224 \\ \hline
            \multicolumn{1}{|l|}{Number of heads}           &    12   \\ \hline
            \multicolumn{1}{|l|}{Number of layers}          &    12   \\ \hline
            \multicolumn{1}{|l|}{Hidden size}               &   768      \\ \hline
            \multicolumn{1}{|l|}{Intermediate size}         &   3072    \\ \hline
            \multicolumn{1}{|l|}{Projection size}         &  768     \\ \hline
            \multicolumn{1}{|l|}{Patch size}               & 16       \\ \hline
            \multicolumn{1}{|l|}{Image size}               & 224   \\ \hline
            \multicolumn{1}{|l|}{Masking ratio}               &   0.5    \\ \hline
            \multicolumn{2}{|c|}{\textbf{Time Series Encoder}}    \\ \hline
            \multicolumn{1}{|l|}{Number of heads}           &    4   \\ \hline
            \multicolumn{1}{|l|}{Number of layers}          &    3   \\ \hline    
            \multicolumn{1}{|l|}{Hidden size}               &   256      \\ \hline
            \multicolumn{1}{|l|}{Intermediate size}         &   3072    \\ \hline
            \multicolumn{1}{|l|}{Projection size}         &  768     \\ \hline
            \multicolumn{1}{|l|}{Positional encoder}    &  \textit{learnable}      \\ \hline
            \multicolumn{1}{|l|}{Normalization}         &  LayerNorm     \\ \hline
            \multicolumn{1}{|l|}{Masking ratio}               &   0.15    \\ \hline
            \multicolumn{1}{|l|}{Average mask length}               &  3    \\ \hline
            \multicolumn{1}{|l|}{Masking sampling strategy}               &     Geometric  \\ \hline
            \multicolumn{1}{|l|}{Time series length}     &  48     \\ \hline
            \multicolumn{2}{|c|}{\textbf{Multimodal Encoder}}     \\ \hline
            \multicolumn{1}{|l|}{Number of heads}           &    12   \\ \hline
            \multicolumn{1}{|l|}{Number of layers}          &    6   \\ \hline
            \multicolumn{1}{|l|}{Intermediate size}         &   3072    \\ \hline
            \multicolumn{1}{|l|}{Projection hidden dimension}  &   2048     \\ \hline
            \multicolumn{1}{|l|}{Projection size}         &  768     \\ \hline
            \multicolumn{2}{|c|}{\textbf{Pretraining}}     \\ \hline
            \multicolumn{1}{|l|}{Learning rate}       &  0.0001      \\ \hline
            \multicolumn{1}{|l|}{Batch size}       &  128     \\ \hline
            \multicolumn{1}{|l|}{AdamW weight decay}       &  0.02     \\ \hline
            \multicolumn{1}{|l|}{AdamW $\beta_1$}       &   0.9     \\ \hline
            \multicolumn{1}{|l|}{AdamW $\beta_2$}       &    0.999   \\ \hline
            \multicolumn{1}{|l|}{Learning rate schedule}       &    Cosine Annealing   \\ \hline
            \multicolumn{1}{|l|}{$\lambda_1$}       &  1.     \\ \hline
            \multicolumn{1}{|l|}{$\lambda_2$}       &  1.    \\ \hline
            \multicolumn{1}{|l|}{$\lambda_3$}       &  0.     \\ \hline
            \multicolumn{1}{|l|}{$\lambda_4$}       &  0.1     \\ \hline
            \multicolumn{1}{|l|}{$\lambda_5$}       &  0.5     \\ \hline
            \multicolumn{1}{|l|}{Total \# of parameters}       &    277.16   \\ \hline
            \multicolumn{2}{|c|}{\textbf{Fine-tuning}}     \\ \hline
            \multicolumn{1}{|l|}{Learning rate}       &  0.0001      \\ \hline
            \multicolumn{1}{|l|}{Batch size}       &  512     \\ \hline
            \multicolumn{1}{|l|}{AdamW weight decay}       &  0.02     \\ \hline
            \multicolumn{1}{|l|}{AdamW $\beta_1$}       &   0.9     \\ \hline
            \multicolumn{1}{|l|}{AdamW $\beta_2$}       &    0.999   \\ \hline
            \multicolumn{1}{|l|}{Learning rate schedule}       &    Cosine Annealing   \\ \hline
            \multicolumn{1}{|l|}{Total \# of parameters}       &    241.62  \\ \hline
            \multicolumn{1}{|l|}{Total \# of trainable parameters}       &    45.54  \\ \hline
\end{tabular}
\end{minipage}%
    \begin{minipage}{.5\linewidth}
    \centering
    \tiny
    \caption*{\small{Amazon Review dataset}}
        \begin{tabular}{|ll|}
            \hline
            \multicolumn{1}{|l|}{Hyper-parameter} & Value \\ \hline
            \multicolumn{2}{|c|}{\textbf{Text Encoder}}           \\ \hline
            \multicolumn{1}{|l|}{HuggingFace model name}   & bert-base-uncased   \\ \hline
            \multicolumn{1}{|l|}{Number of heads}           &    12   \\ \hline
            \multicolumn{1}{|l|}{Number of layers}          &    12   \\ \hline
            \multicolumn{1}{|l|}{Hidden size}               &  768      \\ \hline
            \multicolumn{1}{|l|}{Intermediate size}         &   3072    \\ \hline
            \multicolumn{1}{|l|}{Projection size}         &  768     \\ \hline
            \multicolumn{1}{|l|}{Vocab size}         &  30522   \\ \hline
            \multicolumn{1}{|l|}{Maximum sequence length}         &  512   \\ \hline
            \multicolumn{1}{|l|}{Masking ratio}               &   0.15    \\ \hline
            \multicolumn{2}{|c|}{\textbf{Image Encoder}}          \\ \hline
            \multicolumn{1}{|l|}{HuggingFace model name}           &   google/vit-base-patch16-224 \\ \hline
            \multicolumn{1}{|l|}{Number of heads}           &    12   \\ \hline
            \multicolumn{1}{|l|}{Number of layers}          &    12   \\ \hline
            \multicolumn{1}{|l|}{Hidden size}               &   768      \\ \hline
            \multicolumn{1}{|l|}{Intermediate size}         &   3072    \\ \hline
            \multicolumn{1}{|l|}{Projection size}         &   768     \\ \hline
            \multicolumn{1}{|l|}{Patch size}               & 16       \\ \hline
            \multicolumn{1}{|l|}{Image size}               & 224   \\ \hline
            \multicolumn{1}{|l|}{Masking ratio}               &   0.5    \\ \hline
            \multicolumn{2}{|c|}{\textbf{Tabular Encoder}}    \\ \hline
            \multicolumn{1}{|l|}{Number of heads}           &    4   \\ \hline
            \multicolumn{1}{|l|}{Number of layers}          &    3   \\ \hline    
            \multicolumn{1}{|l|}{Hidden size}               &   1024      \\ \hline
            \multicolumn{1}{|l|}{Attention size in TabNet}         &   64    \\ \hline
            \multicolumn{1}{|l|}{Masking function in TabNet}         &   Sparsemax    \\ \hline
            \multicolumn{1}{|l|}{Projection size}         &  256     \\ \hline
            \multicolumn{1}{|l|}{Masking ratio}               &   0.15    \\ \hline
            \multicolumn{2}{|c|}{\textbf{Multimodal Encoder}}     \\ \hline
            \multicolumn{1}{|l|}{Number of heads}           &    12   \\ \hline
            \multicolumn{1}{|l|}{Number of layers}          &    6   \\ \hline
            \multicolumn{1}{|l|}{Intermediate size}         &   3072    \\ \hline
            \multicolumn{1}{|l|}{Projection hidden dimension}  &   2048     \\ \hline
            \multicolumn{1}{|l|}{Projection size}         &  768     \\ \hline
            \multicolumn{2}{|c|}{\textbf{Pretraining}}     \\ \hline
            \multicolumn{1}{|l|}{Learning rate}       &  0.0001      \\ \hline
            \multicolumn{1}{|l|}{Batch size}       &  64     \\ \hline
            \multicolumn{1}{|l|}{AdamW weight decay}       &  0.02     \\ \hline
            \multicolumn{1}{|l|}{AdamW $\beta_1$}       &   0.9     \\ \hline
            \multicolumn{1}{|l|}{AdamW $\beta_2$}       &    0.999   \\ \hline
            \multicolumn{1}{|l|}{Learning rate schedule}       &    Cosine Annealing   \\ \hline
            \multicolumn{1}{|l|}{$\lambda_1$}       &  1.     \\ \hline
            \multicolumn{1}{|l|}{$\lambda_2$}       &  1.    \\ \hline
            \multicolumn{1}{|l|}{$\lambda_3$}       &  0.01     \\ \hline
            \multicolumn{1}{|l|}{$\lambda_4$}       &  0.     \\ \hline
            \multicolumn{1}{|l|}{$\lambda_5$}       &  0.5     \\ \hline
            \multicolumn{1}{|l|}{Total \# of parameters}       &  288.66M     \\ \hline
            \multicolumn{2}{|c|}{\textbf{Fine-tuning on \textit{Fashion}}}     \\ \hline
            \multicolumn{1}{|l|}{Learning rate}       &  0.00005      \\ \hline
            \multicolumn{1}{|l|}{Batch size}       &  32     \\ \hline
            \multicolumn{1}{|l|}{AdamW weight decay}       &  0.1     \\ \hline
            \multicolumn{1}{|l|}{AdamW $\beta_1$}       &   0.9     \\ \hline
            \multicolumn{1}{|l|}{AdamW $\beta_2$}       &    0.999   \\ \hline
            \multicolumn{1}{|l|}{Learning rate schedule}       &    Cosine Annealing   \\ \hline
            \multicolumn{2}{|c|}{\textbf{Fine-tuning on \textit{Beauty}}}     \\ \hline
            \multicolumn{1}{|l|}{Learning rate}       &  0.0001      \\ \hline
            \multicolumn{1}{|l|}{Batch size}       &  128     \\ \hline
            \multicolumn{1}{|l|}{AdamW weight decay}       &  0.1     \\ \hline
            \multicolumn{1}{|l|}{AdamW $\beta_1$}       &   0.9     \\ \hline
            \multicolumn{1}{|l|}{AdamW $\beta_2$}       &    0.999   \\ \hline
            \multicolumn{1}{|l|}{Learning rate schedule}       &    Cosine Annealing   \\ \hline            
            \multicolumn{1}{|l|}{Total \# of parameters}       &    242.13  \\ \hline
            \multicolumn{1}{|l|}{Total \# of trainable parameters}       &    45.54  \\ \hline            
        \end{tabular}    
    \end{minipage}     
\end{table*}

\section{Architecture Details}
\label{sec:arch}

\noindent\textbf{Text encoder.} We adopt the BERT~\cite{bert} architecture for the text encoder which transforms a tokenized input text into a list of hidden state vectors {$\mathbf{h}_T$}, each corresponding to a tokenized word plus an additional $\mathbf{h}_{CLS,T}$ for the text classification [\texttt{CLS\_T}] token. 

\noindent\textbf{Image encoder.} We use the ViT-B/16~\cite{vit} architecture for the image encoder which receives  images that are divided into patches of size 16 along with positional embeddings and an extra image classification token  [\texttt{CLS\_I}] and encodes them into a list of hidden state vectors {$\mathbf{h}_I$} where each item in the list corresponds to an image patch followed by an additional $\mathbf{h}_{CLS,I}$ for [\texttt{CLS\_I}].

\noindent\textbf{Tabular encoder.} We use TabNet~\cite{tabnet} for encoding tabular (time invariant) features which are represented with numerical values or categorical features. TabNet is an encoder-decoder architecture which encodes tabular data in consecutive multi-steps where each step consists of three processes. First, features are passed into a batch normalization layer followed by a feature Transformer which consists of four gated linear unit (GLU) decision blocks. A split block then divides the processed information to be consumed by an attentive Transformer which performs the sparse feature selection mechanism by learning a mask over salient features. The output for the TabNet encoder is also a list of hidden state vectors generated at the end of each step.

\noindent \textbf{Time series encoder.} We use a conventional Transformer architecture~\cite{transformer} similar to~\cite{mvts} to encode a multivariate time series of a fixed length and certain number of variables. An important consideration regarding time series data is extracting the temporal information effectively. While the positional encodings can preserve some ordering information, the nature of the permutation-invariant self-attention mechanism inevitably results in temporal information loss~\cite{zeng2022transformers}. Therefore, instead of the fixed sinusoidal encoding~\cite{transformer}, we use fully-learnable positional encodings. Similar to all other encoders, the output is a list of hidden states vectors.

\section{\lan's Algorithm}

\begin{algorithm}[t]
\footnotesize
  \caption{Pretraining \lan \label{alg:lanistr}}
    \begin{algorithmic}[1]
      \STATE {{\bfseries Inputs} \lan model weights, Unlabeled parallel and non-parallel data, all hyper-parameters for \lan shown in Table~\ref{tab:hyperparams}}
      \FOR {$epoch=1$ to total number of epochs}
      \STATE {Compute $\mathcal{L}_{\text{MLM}}$ by performing masked language modeling for the text encoder and its decoder}
      \STATE {Compute $\mathcal{L}_{\text{MIM}}$ by performing masked image modeling for the image encoder and its decoder}
      \STATE {Compute $\mathcal{L}_{\text{MFM}}$ by performing masked feature modeling for the tabular encoder and its decoder}
      \STATE {Compute $\mathcal{L}_{\text{MTM}}$ by performing masked time series modeling for the time series encoder (this encoder does not have a decoder)}
      \STATE {Compute $\mathcal{L}_{\text{SimMMM}}$ by performing similarity-based multimodal masking modeling using all the unimodal encoders and the multimodal encoder-decoder module}
      \STATE {Compute $\mathcal{L}_{\text{\lan}}$ by combining all the pretraining objectives as shown in Eq. \ref{eq:total_loss}}
      \STATE Perform back-propagation and update \lan's weights using the total loss in $\mathcal{L}_{\text{\lan}}$.
      \ENDFOR
    \end{algorithmic}
    \end{algorithm}

\section{Experimental Details}

\subsection{Datasets licenses}\label{sec:license}
We use two publicly-available datasets to construct our benchmarks. These datasets can be downloaded from their original hosts under their terms and conditions. For MIMIC-IV dataset, Only credentialed users who sign the data use agreement can access the files and there is a training required to use the data in research. 

\begin{itemize}
    \item MIMIC-IV v2.2~\cite{mimic4} License can be found at  \url{https://physionet.org/content/mimiciv/view-license/2.2/} 
    and instructions to download and term of use can be found at  \url{https://physionet.org/content/mimiciv/2.2/}.
    \item Amazon Review Data (2018)~\cite{apr} License, instructions to download, and term of use can be found at \url{https://nijianmo.github.io/amazon/index.html} 
\end{itemize}

\subsection{Preprocessing structured data}
For time-series sequences in MIMIC-IV, similar to~\cite{medfuse,harutyunyan2019multitask} we use 17 clinical variables from which five are categorical (capillary refill rate, Glasgow coma scale eye opening, Glasgow coma scale motor response, Glasgow coma scale verbal response, and Glasgow coma scale total) and 12 are continuous (diastolic blood pressure, fraction of inspired oxygen, glucose, heart rate, height, mean blood pressure, oxygen saturation, respiratory rate, systolic blood pressure, temperature, weight, and pH). We regularly sample the input every one hour over the course of 48 hours, discretize and standardize the clinical variables to obtain the input. After pre-processing and one-hot encoding of the categorical features, we obtain a vector representation of size 48 at each time step. 

For tabular data in Amazon Review dataset, we also use one-hot encoding for categorical features and fill missing values with the mean of that columns.

\subsection{Datasets details}
\noindent \textbf{MIMIC-IV}. In total we used  3,680,784 samples for pretraining and all hyperparameters used in our experiments are shown in Table~\ref{tab:hyperparams} on this dataset which is constructed using 377,110 images, 331,794 notes, and 25,071 time series for different stays in the hospital. For fine-tuning, we split the labeled parallel samples randomly such that there is no overlap in stays for the same patient in train/validation/test splits. This results in 5797 parallel samples for training while 54 and 617 samples were used for validation set and test set, respectively. The validation set was mainly used to tune fine-tuning hyper-parameters including learning rate, batch size, and weight decay. 

\noindent \textbf{Amazon Review Dataset}. In total we used $\text{5,581,312}$ non-parallel samples from \textit{Office Products} category for pretraining. For fine-tuning, we used 512, 128, and 256 labeled parallel samples for train, validation, and test sets, respectively. We used the validation set for tuning the hyper-parameters including learning rate, batch size, and weight decay.

\subsection{Compute}
On MIMIC-IV dataset we used 8$\times$A100 40GB-SXM4 NVIDIA GPUs for both pretraining and fine-tuning stages. Total wall-clock time for pretraining is 280 hours (40 epochs) and for fine-tuning is 576 minutes (500 epochs). For Amazon Review dataset, we used 16$\times$A100 40GB-SXM4 NVIDIA GPUs for pretraining and 8 GPUs for fine-tuning which took 130 hours (20 epochs) and 9 minutes (200 epochs) of wall-clock time, respectively.


\section{Limitations}\label{sec:limitations}
In this work, we evaluate \lan on datasets with three modalities (image+text+tabular) or (image+text+time series), although our framework can be extended to four modalities altogether. In our current version of \lan, we do not have a mechanism to determine the effectiveness of training with all the modalities in hand prior to initiating the experiments. For instance, the MIMIC-IV dataset also provides tabular data, which contains the demographic information of patients, such as gender, marital status, insurance company, age, and so on. However, we find that using all four modalities (image+text+time series+tabular) yields similar performance to using image+text+time series only. Therefore, we omit the tabular modality. While this might have an intuitive explanation that demographic information can be irrelevant to our studied downstream task, which is mortality prediction within 48 hours of ICU stay, it is still desirable to develop an automated prediction tool to determine modality importance prior to the fine-tuning stage.
On the other hand, extensions to support other modalities like audio and video, would be important future directions. 

\end{document}